%% file: acl_latex.tex
\pdfoutput=1

\documentclass[11pt]{article}

\usepackage[]{acl}

\usepackage{times}
\usepackage{latexsym}
\usepackage{amssymb}

\usepackage[T1]{fontenc}

\usepackage[utf8]{inputenc}
\usepackage{fullpage}
\usepackage[T1]{fontenc} 
\usepackage{epsf} 
\usepackage{graphics} 
\usepackage{psfrag,xspace}
\usepackage{color,etoolbox}
\usepackage[ruled,vlined]{algorithm2e}
\usepackage{mathtools}  
\usepackage{thm-restate}
\usepackage{dsfont}
\usepackage{booktabs} 
\usepackage{microtype}

\usepackage{inconsolata}
\usepackage{comment}

\usepackage{xcolor}
\definecolor{oceanblue}{rgb}{0.4, 0.7, 0.85} 
\definecolor{sunsetorange}{rgb}{0.95, 0.65, 0.4} 
\definecolor{mintgreen}{rgb}{0.8, 1.0, 0.8} 
\definecolor{lavenderpurple}{rgb}{0.85, 0.75, 1.0} 

\usepackage{colortbl}

\newcommand{\ourmodelshort}{\texttt{OrchestraLLM}\xspace}
\newcommand{\SLMshort}{Prompt-DST\xspace}
\newcommand{\LLMshort}{IC-DST\xspace}
\input{Definitions}
\title{OrchestraLLM: Efficient Orchestration of Language Models for Dialogue State Tracking}

 \author{Chia-Hsuan Lee \\
 University of Washington\\
   \texttt{chiahlee@uw.edu} \\\And
   Hao Cheng \\
   Microsoft Research \\
   \texttt{chehao@microsoft.com}\\\And
   Mari Ostendorf\\
   University of Washington\\
   \texttt{ostendor@uw.edu
}}

\begin{document}
\maketitle
\begin{abstract}
\input{sections/abstract2}

\end{abstract}

\input{sections/intro3}
\input{sections/dst_models}
\input{sections/router2}
\input{sections/experiment}

\input{sections/analysis}
\input{sections/related2}
\input{sections/conclusion}
\input{sections/limitations}

\bibliography{anthology,custom}
\pagebreak
\appendix
\input{sections/appendix}

\end{document}

%% file: sections/abstract2.tex
Large language models (LLMs) have revolutionized the landscape of Natural Language Processing, but are computationally expensive. To reduce the cost without sacrificing performance, previous studies have explored various approaches to harness the potential of Smaller Language Models (SLMs) as cost-effective alternatives to their larger counterparts. 
Driven by findings that SLMs and LLMs exhibit complementary strengths in a structured knowledge extraction task, this work presents a novel SLM/LLM routing framework designed to improve computational efficiency and enhance task performance. In dialogue state tracking tasks, the proposed routing framework enhances performance substantially compared to relying solely on LLMs, while reducing the computational costs by over 50\%.

%% file: sections/intro3.tex
\section{Introduction}
Large Language Models (LLMs) have become versatile tools capable of tackling a wide range of tasks with only a few training examples. However, their expanding sizes have brought escalating computational demands. In contrast, more efficient Smaller Language Models (SLMs) often require a substantial amount of fine-tuning data to become truly effective. This work addresses scenarios where only limited task-specific data is available, making fine-tuned SLMs less dependable. Our objective is to develop a routing framework that orchestrates SLMs and LLMs, enhancing task performance while reducing computational costs.

Task-oriented dialogue is crucial for efficient human-computer interaction, enabling systems to understand and assist with specific tasks like booking flights or scheduling meetings. Task-oriented dialogues involving structured data typically rely on Dialogue State Tracking (DST), where user intent is extracted from the dialogue history between a user and the agent in the form of slot values associated with a predefined schema. Fine-tuned SLMs have been used in DST for a few years, including both autoregressive LMs ~\cite{ham2020end,hosseini2020simple,peng2020soloist} and sequence-to-sequence LMs~\cite{lee2021dialogue,su2022multi,bang-etal-2023-task,imrattanatrai-fukuda-2023-end,wang2023divide}. LLMs have been used for few-shot in-context learning in DST~\cite{xie2022unifiedskg,hudevcek2023large,hu-etal-2022-context,king2023diverse} where LLMs  are prompted with human-authored task descriptions or in-context exemplars. In our work, we seek to take advantage of the effectiveness of LLMs with a small amount of training data but reduce the cost.

Strategies that leverage both SLMs and LLMs have been developed to mitigate the computational demands of LLMs. Cascade-based approaches  direct a query to an LLM when it cannot be resolved by an SLM \cite{chen2023frugalgpt, madaan2023automix}. These approaches introduce latency and computational redundancy since they consistently query SLMs. Other approaches use binary classifiers to predict the most appropriate LM to utilize \cite{kag2022efficient, vsakota2023fly}. A limitation of the classifier-based approaches is the necessity for retraining when introducing new models.

In this work, we propose a dynamic routing framework, \textbf{\ourmodelshort} (illustrated in \autoref{fig:system}), that leverages small (fine-tuned) and large LM experts.
Hypothesizing that examples with similar semantic embeddings are of the same difficulty level, we select an appropriate expert based on embedding distances between the testing instance and instances in expert pools. The expert pools contain examples representing the types of dialogue contexts where the different LMs provide more reliable answers. 
After retrieving the top k nearest examples, an expert is selected based on the majority vote. Unlike cascade-based and classifier-based approaches, the proposed framework eliminates the need for router training, though hand-labeled data is needed for creating the expert pools. In addition, the retriever can be fine-tuned with target task labels or expert information to achieve more efficient and accurate routing.

In summary, the key contribution of this work is the introduction of a novel switching model designed to reduce the computational costs associated with LLMs while simultaneously enhancing performance. Experimental results on two different multi-domain DST benchmarks (MultiWOZ~\cite{budzianowski2018multiwoz,ye-etal-2022-multiwoz} and SGD~\cite{rastogi2020towards}) demonstrate that \ourmodelshort capitalizes on the proficiencies of different experts, outperforming LLM systems while also achieving a substantial reduction of over 50\% in computational costs.


%% file: sections/dst_models.tex
\section{Dialogue State Tracking}
In this work, we focus on combining general-purpose LLMs and task-specific SLMs to achieve better efficiency for dialogue state tracking (DST).
In the following, we first provide the necessary task setups and then detail the two representative DST models using LLMs and SLMs respectively.

A task-oriented dialogue (TOD) consists of a sequence of exchanges between two parties, each of which is initialized by the user and followed by a response from the system. 
Here, we denote each exchange as a turn leading to a sequence, $U_{1}, A_{1}, ..., U_{T}, A_{T}$, where $U_t$ and $A_t$ represent the user utterance and the system response, respectively.
For the $t$-th turn, the user provides a new utterance \textit{$U_{t}$}, and the system agent responds with utterance \textit{$A_{t}$}.
At turn $t$, the corresponding dialogue context is $C_t=\{U_{1}, A_{1}, \ldots, A_{t-1}, U_{t}\}$, which excludes the latest system response $A_t$.
The goal of DST is to extract task-relevant information as structured representations (dialogue states) from user-system utterances so that the user requests can be fulfilled accordingly.
To facilitate this, there is typically a task-specific schema provided.
In a multi-domain scenario considered in this paper, the schema contains $M$ domains $\mathcal{D}=\{d_1, \ldots, d_M\}$ and
$N$ slots $\mathcal{S}=\{s_1, \ldots, s_N\}$ to track.
$DST_t$, the dialogue state at turn $t$, defines the current mappings from pairs of ($d_m$, $s_n$) into a value $v$ based on dialogue context $C_t$.
Specifically, 
\begin{align*}
    DST_t = \{ (d_m, s_n, v_{mn}^t)| v_{ij}^t\neq \texttt{null} \},
\end{align*}
only containing the non-null slots accumulated so far.
Instead of directly predicting the entire dialogue state from scratch,
we build dialogue state predictions based on the turn-level belief (TLB) as done by \citet{hu-etal-2022-context}, which allows a more flexible combination of LLMs and SLMs.
At turn $t$, the DST model only predicts $TLB_t$, where new expressed slots or slots with updated values are used to get the latest $DST_t$ via aggregating all previous TLBs.\footnote{We replace previous values with updated ones for slots present in prior TLBs.} 

In the literature, task-specific SLM-based DST models are typically fine-tuned with full-parameter updates while DST models using LLMs are realized via few-shot in-context learning.
We discuss the two different DST models considered below. 
\paragraph{LLM DST.}
\label{sec:IC-DST}
\textit{IC-DST} \cite{hu-etal-2022-context} is an in-context learning (ICL) framework that enables few-shot DST with LLMs.
The prediction is the change in each turn pair instead of the accumulated dialogue states.
To obtain the accumulated dialogue states, the turn changes are aggregated across turns. Dialogue states of previous turns are used as a summary of the context history and it allows to fit in more exemplars which is crucial for ICL performance.
Concretely, given the schema table, $K$ in-context exemplars, dialogue states of the previous turn, and the input instance (most recent agent-user utterance pair), the LLM outputs
\begin{equation}
    TLB_t = \text{LLM} (T, E_{1:K}, DST_{t-1}, A_{t-1}, U_t)
\end{equation}
where $T$ is the schema table for all domains, $E_k$ are examples of pairs of turn changes and associated outputs.

\paragraph{SLM DST.}
\label{sec:SDP-DST}
Here, we develop a prompt-based DST model (denoted as \textit{Prompt-DST}) with SLM (T5~\cite{raffel2020exploring}). The input of Prompt-DST is similar to IC-DST, except that the in-context exemplars are excluded. Specifically, given the schema prompt-augmented input, the model outputs
\begin{equation}
    TLB_t = \text{SLM} (T, DST_{t-1}, A_{t-1}, U_t).
\end{equation}
Here, the model is trained using the learning objective by maximizing the log-likelihood of slot values $v_t(d_m, s_n)$ for the current TLB, \ie
\begin{equation}
\max \log P(TLB_t| T, DST_{t-1}, A_{t-1}, U_t).
\end{equation}
During inference, a greedy decoding procedure is directly applied, \ie only the most likely token in the given model vocabulary is predicted at each decoding step.

\begin{figure*}[t]
  \includegraphics[width=1.0\textwidth]{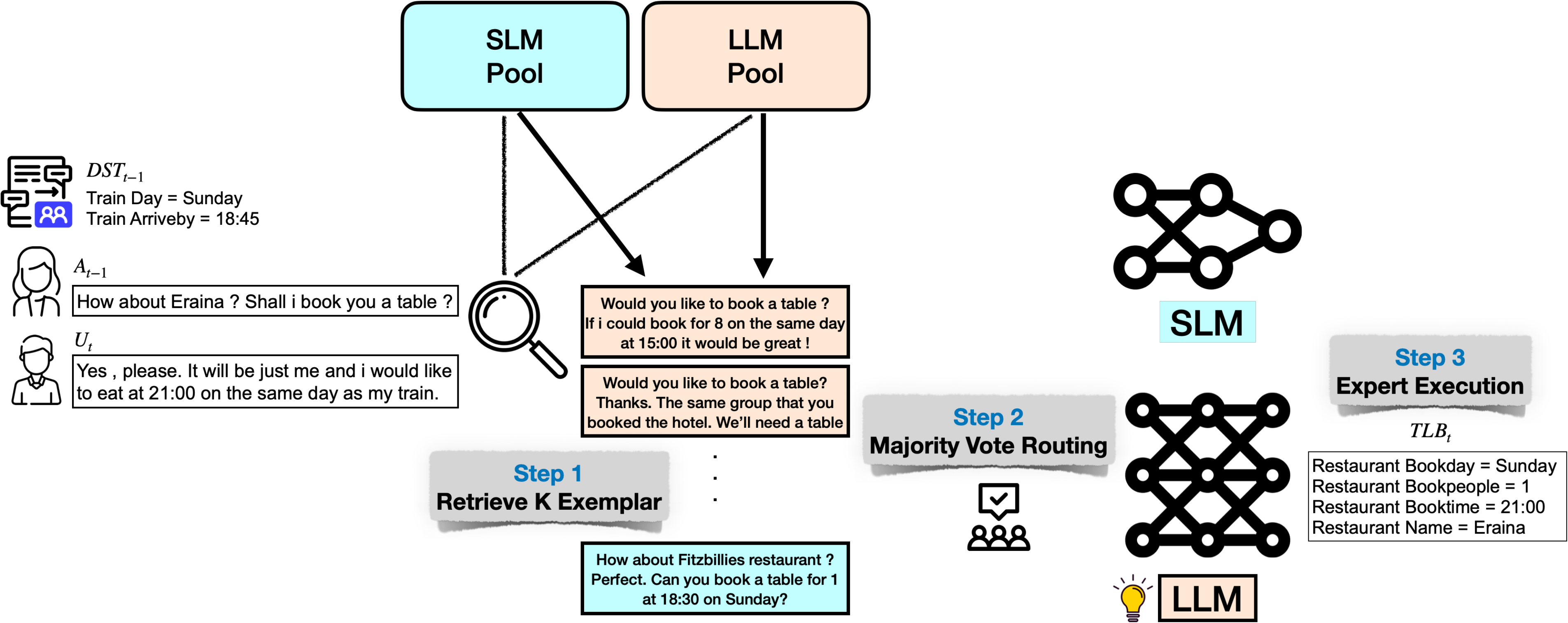}
  \caption{
Illustration of \ourmodelshort. LMs are orchestrated by a retrieval-based dynamic router. During inference, the testing instance queries the expert pools to retrieve top k similar examples. Subsequently, a LM expert is selected based on the majority vote.} 
  \label{fig:system}
\end{figure*}

%% file: sections/router2.tex
\section{Routing Approach}
Here, we present our approach for routing with \ourmodelshort applied to the DST task. 
The overall framework is illustrated in \autoref{fig:system}.
We denote different DST models as \textbf{experts}.
Given a new input instance (the triplet $(DST_{t-1}, A_{t-1}, U_t)$), \ourmodelshort first computes its semantic embedding, compares it with exemplar embeddings of triplets from each \textbf{expert pool} using a cosine distance, and retrieves the top-K exemplars. The router assigns the input to an expert based on majority vote. While our approach draws inspiration from the work of ~\citet{jang2023exploring}, it is important to note that their approach primarily focuses on optimizing task performance in zero-shot task transfer scenarios, whereas our emphasis lies in improving computational efficiency within the few-shot learning settings.


\subsection{Expert Pool Construction}
For each dialogue in a small held-out set, the SLM and LLM experts are used to predict the TLB at each user turn ($TLB_t$) individually. 
If both experts correctly predict the TLB, the instance triplet is included in the SLM pool. When only one expert correctly predicts the TLB, the instance is assigned to that expert's pool. Instances that are not correctly predicted are not used in either pool.

\subsection{Triplet Representation Learning}
Similar to recent work on dense retrieval \cite{karpukhin-etal-2020-dense},
the retriever uses a bi-encoder architecture, which encodes dialogues with labels and predictions into embedding space.
Throughout the work, SenBERT \cite{reimers-2019-sentence-bert} is used as the backbone embedding model.
The bi-encoder is fine-tuned using a small set of dialogues, the same as that used to construct the expert pools.
We use a contrastive loss such that the similarity between a positive example pair is high and the similarity between a negative example pair is low. 
Three different methods for constructing positive and negative pairs are explored: task-aware, expert-aware, and their combination.

{\bf Task-Aware Supervision} identifies
positive and negative instance pairs for training by first computing pairwise similarity for each sample in the hold-out set. 
Then, the $l$ highest and lowest scoring pairs are used as positive and negative examples, respectively.
The similarity function leverages the gold annotations of the hold-out set dialogues. Given two instances, $a$ and $b$, the similarity is a weighted combination of the slot-value similarity of the previous state (DST) and the current TLB:
\begin{align*}
    \frac{1}{2} Sim_{\mathit{DST}} + Sim_{\mathit{TLB}} .
\end{align*}
Let 
$TLB_\mathit{x}=\{(s_1^x, v_1^x), \cdots, (s_m^x, v_m^x)\}$ be the TLB of instance $x$.
Following \citet{hu-etal-2022-context}, the slot-value pair similarity is 
\begin{align*}
    F_{\mathit{slot\mbox{-}value}} = F(&\{(s_1^a, v_1^a), \cdots,(s_m^a, v_m^a)\},\\
    &\{(s_1^b, v_1^b), \cdots, (s_n^b, v_n^b)\}) .
\end{align*} 
and the slot similarity is
\begin{align*}
    F_{\mathit{slot}} = F(\{s_1^a, \cdots,s_m^a\}, \{s_1^b, \cdots, s_n^b\}).
\end{align*}
where $F$ is the standard definition of F1 score \ie $F=\frac{2PR}{P+R}$, in which $P$ is precision, and $R$ is recall. 
The similarity score between $TLB_\mathit{a}$ and $TLB_\mathit{b}$ is 
\begin{align*}
Sim(TLB_\mathit{a}, TLB_\mathit{b}) = F_{\mathit{slot\mbox{-}value}} + F_{\mathit{slot}} - 1
\end{align*}
The context history similarity $Sim_{\mathit{DST}}$ is defined in the same way.

{\bf Expert-Aware Supervision} first groups instances in the hold-out set according to which expert gave the most accurate prediction. (For ties, the SLM is chosen.)
We then compute pairwise triplet similarities using an off-the-shelf embedder (\eg SenBERT). The $l$ highest scoring pairs with the same expert label are positive examples, and the $l$ lowest scoring pairs with different expert label are negative examples.

{\bf Task+Expert-Aware Supervision} simply pools both sets of positive and negative pairs.

Note that task-aware supervision is agnostic to what experts are used in routing, so the embedding model need not be retrained as experts are added or updated. Expert-aware supervision will require updating the embedding model if the experts change.
In all cases, the expert pools will need to be updated with changes to the experts. 

%% file: sections/experiment.tex
\section{Experiments}
\subsection{Datasets}
We use two datasets detailed below for experiments.
A summary of DST datasets is reported in \autoref{tab:data_sum}.
\input{tables/datasets}

\noindent
\textbf{MultiWOZ} \cite{budzianowski2018multiwoz} is a multi-domain task-oriented dialogue dataset that contains over 10K human-human written dialogues across 8 domains and has been one of the most popular benchmarks in the DST literature.
After the publication of \citet{budzianowski2018multiwoz}, many works improve the label qualities, \eg MultiWOZ 2.1 \cite{eric2020multiwoz} and MultiWOZ 2.4 \cite{ye2021multiwoz}.
We experiment using the most recent version, MultiWOZ 2.4.

\noindent
\textbf{SGD \cite{rastogi2020towards}} 
 is a task-oriented dialogue dataset that contains over 16k multi-domain conversations spanning 41 domains, featuring out-of-domain evaluation. 15 out of 21 domains in the test set are not present in the training set and 77\% of the dialogue turns in the test set contain at least one domain not present in the training set. 

\subsection{Experimental Setting}
In this work, we consider a \textbf{few-shot} set up for DST.
Following the multi-domain experiment setting from ~\citet{wu2020improving}, we randomly sample 5\% of training data from MultiWOZ and SGD respectively for training the expert models. 

\paragraph{Model and Hyperparameter Setting.}
For \SLMshort, we use T5-base and T5-large as the backbone model for MWOZ and SGD respectively, as the latter is more complex in terms of schema and more dialogue turns. For IC-DST, we use ChatGPT as the backbone model\footnote{Accessed: August–October 2023, Version: gpt-3.5-turbo-0301.} with 10 in-context exemplars.
We initialize the routing retriever from SenBERT (all-mpnet-base-v2).
We run inference on 100 dialogues randomly sampled from validation sets of MWOZ and SGD as the held-out sets. The same 100 dialogues are used to train the retriever.
For all experiments, $l=25$ is used for the positive and negative examples for contrastive learning. During inference, we randomly sample 100 turns from the held-out sets to serve as SLM pool and LLM pool respectively for MWOZ experiments and 300 turns for SGD experiments.\footnote{We also experimented with 50 turns for MWOZ and observed less than 1\% accuracy degradation.} We leave it for future work to create novel strategies for expert pool instance selection. We use $k=10$ for the majority vote and break the tie by favoring SLM.

\input{tables/dualmodel_mwoz}
\input{tables/dualmodel_sgd}
\subsection{Evaluation}
\subsubsection{Accuracy}
Conventionally, DST systems are evaluated by joint goal accuracy (JGA) on accumulated dialogue states~\cite{henderson-etal-2014-second}. This metric assesses the correctness of the dialogue state at each turn and deems it accurate only if all slot values within every domain precisely match the ground-truth values. It is difficult to accurately assess how well a system performs on single turns with DST JGA. Therefore we also report turn-level belief (TLB JGA)~\cite{dey2022towards}.

\subsubsection{Efficiency}
Floating-point operations per Second (FLOPs) represent the number of floating-point arithmetic operations (additions and multiplications) a model performs in one pass. FLOPs are often used to estimate the computational cost or workload required for training or inference. 
Training a large model requires a significant number of backward passes, which are more expensive than forward passes, yet inference is a continuous process that happens whenever the model is in use, thus accruing more cost over time. NVIDIA~\cite{leopold2019awst4}
and Amazon~\cite{barr2019amazon} report around 90\% of the ML workload is inference processing in their cloud services. Therefore, we choose to report FLOPs for inference time usage.

We estimate the aggregate computational cost, measured in TeraFLOPs, required for performing inference across the entire testing dataset. It is important to note that IC-DST relies on ChatGPT, a model that is not publicly accessible, thus precluding a direct evaluation of its computational efficiency. Based on prevailing conjecture within the public domain, ChatGPT is presumed to be a fine-tuned iteration of the GPT-3 model with a substantial parameter count of 175 billion~\cite{brown2020language}. To estimate the computational requirements, we conduct FLOPs measurements on the GPT-2~\cite{radford2019language} model and subsequently scale these measurements in accordance with the parameter size differential between GPT-2 and ChatGPT. The computational cost of the retriever, measured in FLOPs, for each turn instance, is approximately 0.02 TeraFLOPs. This computational load becomes negligible when considered in conjunction with ChatGPT in the \ourmodelshort. ChatGPT requires approximately 3000 TeraFLOPs for each turn instance.

\subsection{Baselines}
\textbf{Classification-Based Routing}\\
We compare our routing framework with existing classification-based approaches to model switching, such as those proposed by \citet{vsakota2023fly} and \citet{kag2022efficient}. These existing approaches typically train a binary classifier to serve as the router. We train BERT~\cite{devlin-etal-2019-bert} (bert-base-cased) with the expert labels in the hold-out set of dialogues with a binary classification objective to do routing as a baseline.
\textbf{Cascade-Based Routing}\\
Cascade-based approaches~\citet{chen2023frugalgpt,madaan2023automix} typically query a SLM and redirect the instance to a LLM if the smaller language model is not confident enough. We choose to utilize the normalized sequence level probability of SLM output as the confidence measure. We tune the probability threshold on the hold-out-set and use the threshold to determine whether to redirect the instance to LLM during inference.

\subsection{Results}
\subsubsection{MultiWOZ} 
We demonstrate the MultiWOZ experiments in a few-shot setting in \autoref{tab:dualmodel_mwoz}. We use 5\% of dialogues in the training set for finetuning \SLMshort and retriever of \LLMshort. \SLMshort and \LLMshort perform inference on another 100 dialogues from the validation set, documenting the turns each expert specializes in. We randomly select 100 turns from these dialogues for each expert to serve as expert pools for dynamic routing. 

As expected, \LLMshort outperforms \SLMshort in the few-shot setting, indicating that the LLM is more generalizable than the fine-tuned SLM. The BERT-based classification router struggles to effectively harness the capabilities of both models. To establish an upper performance bound for the learned router, we introduce the oracle router, which aggregates predictions from both LLM and SLM when either model is correct, with a preference for SLM whenever available. Even with a vanilla SenBERT as a retriever, \ourmodelshort outperforms \LLMshort while saving 60\% calls to LLM, demonstrating the effectiveness of our proposed framework. Further finetuning the retriever with the proposed task-aware contrastive examples routes examples more effectively and improves DST JGA around 3\% compared to \LLMshort. With additional expert-aware training of the retriever, we can further save around 7\% traffic to LLM with superior performance compared with \LLMshort.
\vspace{-1mm}
In spite of its compact size, \SLMshort is finetuned to align with specific in-domain knowledge and task-specific artifacts (\eg schema constraints and customized labeling strategies). Conversely, \LLMshort is enriched with an extensive repertoire of knowledge acquired during the pretraining phase of LLM, endowing it with contextual reasoning capabilities and an enhanced grasp of common-sense knowledge (Section~\ref{sec:example}). Since these two models are complementary, an effective integration can surpass the performance of the \LLMshort model.

\input{tables/crossdataset_retriever}

\subsubsection{SGD} 
To evaluate our system under out-of-domain scenarios, we show experimental results in a few-shot setting on SGD in \autoref{tab:dualmodel_sgd}. We use 5\% of dialogues in the training set for finetuning \SLMshort and the retriever of \LLMshort. \SLMshort and \LLMshort performed inference on another 100 dialogues in the validation set to serve as expert pool. We randomly select 300 turns from each expert to serve as expert pools for dynamic routing. 

As we observe in MultiWOZ, incorporating an off-the-shelf SenBERT as the router improves the TLB score and also saves around 50\% of computes. Finetuning SenBERT with the task-aware objective improves efficiency by 5\% and increases both the TLB and DST scores. With the additional expert-aware supervision, more turns are routed to SLM and improves TLB score. This setting outperforms \LLMshort by over 4\% TLB JGA and saves 57\% FLOPs, demonstrating that our router is universal enough to support cross-domain assignment and successfully improves system accuracy.

%% file: tables/datasets.tex
\begin{table}[t]
    \small
    \centering
    \begin{tabular}{l@{\hskip3pt}c@{\hskip3pt}c}
    \toprule
        Dataset &  MultiWOZ &  SGD  \\
        \midrule
        \# Domains & 8 & 41  \\
        \# Dialogues & 8438 & 16142 \\
        \# Total Turns & 113556 & 329964   \\
         Avg. Turns per Dial. & 13.46 & 20.44  \\   
         Avg. Toks per Turn & 13.13 & 9.75   \\   
        \# Slots & 24 & 214 \\ 
        \# Slot Values & 4510 & 14139 \\ 
        \bottomrule
    \end{tabular}
    \caption{Experiment data summary. The numbers are computed on training splits of the datasets.
    \label{tab:data_sum}}
\end{table}

%% file: tables/dualmodel_mwoz.tex
\begin{table*}[t]
    \small
    \centering
    \begin{tabular}{l|c|cccc}
    \toprule
        \textbf{Models}  & \textbf{Router}  & \textbf{Assignment Ratio}  & \textbf{TeraFLOPs} & \textbf{TLB JGA} & \textbf{DST JGA} \\
         \midrule
         \rowcolor{lightgray} \multicolumn{6}{c}{\textbf{DST Baselines}} \\
        \SLMshort & N/A & N/A & 272 & 73.43 & 46.06   \\
        \LLMshort~\cite{hu-etal-2022-context} & N/A & N/A & 22 M & 
 78.21 & 49.68 \\
        DS2 - T5~\cite{shin-etal-2022-dialogue}$\ast$ & N/A & N/A & N/A & N/A & 49.89   \\
        \midrule
        \rowcolor{lightgray} \multicolumn{6}{c}{\textbf{Routing Baselines}} \\
        Prompt-DST \& IC-DST  & Oracle & 73\% Prompt-DST & 5.94 M & 88.07 & 65.39 \\
         Prompt-DST \& IC-DST & Classification-Based & 91\% Prompt-DST &  1.98 M & 77.60 & 47.58 \\
        Prompt-DST \& IC-DST & Cascade-Based & 13 \% Prompt-DST &  19.14 M & 80.40 & 51.46 \\
        \midrule
        \rowcolor{mintgreen} \multicolumn{6}{c}{\textbf{Our Retrieval-Based Routing DST}} \\
         \textbf{\ourmodelshort} & SenBERT & 60\% Prompt-DST & 8.8 M & 80.74 & 50.19 \\
          \textbf{\ourmodelshort} & Task-Aware & 55\% Prompt-DST & 9.9 M  & 
         82.43 & 52.53 \\
          \textbf{\ourmodelshort} & Expert-Aware & 78\% Prompt-DST & 4.8 M  & 81.02 & 50.65 \\
          \textbf{\ourmodelshort} & Task+Expert-Aware & 62\% Prompt-DST & 8.3 M  &  \textbf{82.46} &  \textbf{52.68} \\
        \bottomrule
    \end{tabular}
    \caption{Results on MultiWOZ 2.4. The TeraFLOPs are computed on inference passes on the entire testing set. We report the percentage of turns routed to \SLMshort in the assignment ratio column. $\ast$ marks numbers reported in ~\citet{hu-etal-2022-context}.}
    \label{tab:dualmodel_mwoz}
\end{table*}

%% file: tables/dualmodel_sgd.tex
\begin{table*}[t]
    \small
    \centering
    \begin{tabular}{l|c|cccc}
    \toprule
        \textbf{Models}  & \textbf{Router}  & \textbf{Assignment Ratio}  & \textbf{TeraFLOPs} & \textbf{TLB JGA} & \textbf{DST JGA} \\
        \midrule
        \rowcolor{lightgray} \multicolumn{6}{c}{\textbf{DST Baselines}} \\
        \SLMshort & N/A & N/A & 8882 & 62.21 & 28.38   \\
        \LLMshort~\cite{hu-etal-2022-context} & N/A & N/A & 121 M & 
63.86 & 33.15 \\
        \midrule
        \rowcolor{lightgray} \multicolumn{6}{c}{\textbf{Routing Baselines}} \\
        Prompt-DST \& IC-DST & Oracle & 62\% Prompt-DST & 45.98 M & 77.48 & 47.50 \\
        Prompt-DST \& IC-DST & Classification-Based & 38\% Prompt-DST & 75.02 M & 66.94 & 31.86 \\
        Prompt-DST \& IC-DST & Cascade-Based & 7.9\% Prompt-DST & 111.34 M & 64.17 & 32.75 \\
        \midrule
        \rowcolor{mintgreen} \multicolumn{6}{c}{\textbf{Our Retrieval-Based Routing DST}} \\
          \textbf{\ourmodelshort} & SenBERT & 50\% Prompt-DST & 60.50 M & 65.97 & 32.75 \\
          \textbf{\ourmodelshort} & Task-Aware & 55\% Prompt-DST & 54.45 M & 67.25 & 32.78 \\
         \textbf{\ourmodelshort} & Expert-Aware & 54\% Prompt-DST & 55.66 M & 67.34 & 32.95 \\
          \textbf{\ourmodelshort} & Task+Expert-Aware & 57\% Prompt-DST & 52.03 M  & \textbf{68.09} & \textbf{33.07} \\
        \bottomrule
    \end{tabular}
    \caption{Results on SGD. The TeraFLOPs are computed on inference passes on the entire testing set. We report the percentage of turns routed to Prompt-DST in the assignment ratio column.}
    \label{tab:dualmodel_sgd}
\end{table*}

%% file: tables/crossdataset_retriever.tex
\begin{table*}[t]
    \small
    \centering
    \begin{tabular}{l|cc|cc}
    \toprule
        \textbf{Router} & \multicolumn{2}{c}{\textbf{SGD}} & \multicolumn{2}{c}{\textbf{MWOZ}} \\
        & \textbf{\textit{Assignment Ratio}}  & \textbf{TLB JGA}  &\textbf{\textit{Assignment Ratio}}  & \textbf{TLB JGA}  \\
        \midrule
        SenBERT & 50\%  & 65.97 & 60\%  & 80.74   \\
        Task-Aware (SGD) & 55\% & 67.25  & 54\% & 80.75  \\
        Task-Aware (MWOZ) & 43\% & 66.57  & 55\%   & 82.43   \\
        \bottomrule
    \end{tabular}
    \caption{Cross-dataset routing results of \ourmodelshort on SGD and MWOZ. We denote the \% of testing turns routed to Prompt-DST (SLM) as \textit{Assignment Ratio}.}
    \label{tab:crossdataset_retriever}
\end{table*}

%% file: sections/analysis.tex
\section{Analysis}

\subsection{Cross-Domain Generalization}
\begin{figure}[]
 \centering
  \includegraphics[width=\linewidth]{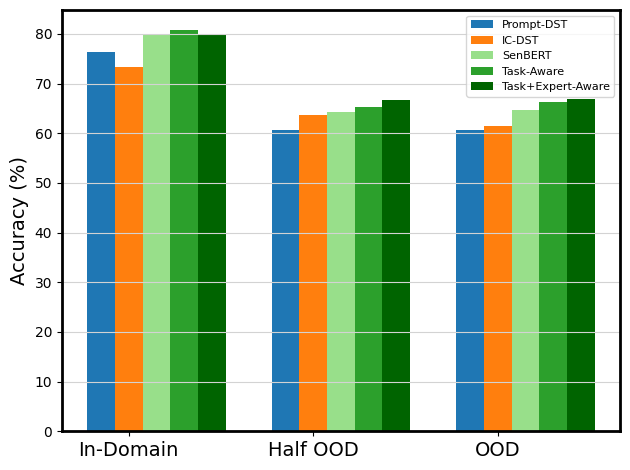}
  \caption{Cross-domain generalization results on SGD. We denote \textit{In-Domain} when all of the testing domains are in the training set and denote \textit{OOD} when all of the testing domains are not in the training set. For all other dialogues, we categorize them as \textit{Half OOD}. We report TLB JGA for all settings. Green bars indicate \ourmodelshort with different retrievers.} 
  \label{fig:OOD_SGD}
\end{figure}

\textbf{Out-of-Domain (OOD) in SGD}
To assess the effectiveness of \ourmodelshort in generalizing to unseen domains, we present breakdown results on SGD in Figure~\ref{fig:OOD_SGD}.
First, we observe that \SLMshort performs better than \LLMshort on in-domain dialogues but lags behind \LLMshort on all other types of dialogues. This suggests that the generalization ability of Large Language Models (LLMs) is superior to Smaller Language Models (SLMs). All variants of \ourmodelshort outperform \LLMshort in OOD scenarios, demonstrating the router's capability to effectively dispatch instances even when they are out of the domain.

\textbf{Cross-Dataset Retriever}
We further evaluate our proposed framework in a more challenging scenario where the router and backbone models are trained in different datasets. We train the retriever model on MWOZ holdout set dialogues and evaluate the framework on SGD testing dialogues and vice versa. 
The results are shown in Table~\ref{tab:crossdataset_retriever}.
Notably, our routing framework can still effectively orchestrate two LLMs with a retriever trained with a different dataset and outperforms \LLMshort while also achieving computational cost savings of approximately 54\% on MultiWOZ and 43\% on SGD. Compared with the dataset-matched finetuning, the mismatched finetuning with a different dataset only slightly hurts the accuracy and efficiency.


\input{tables/example}

\subsection{Specialty of SLM and LLM}
\label{sec:example}
To better understand the complementary nature of the LMs, we inspected examples to
identify their specialties. We provide representative examples from the expert pools in Table~\ref{tab:example}.
One common mistake made by LLM is failing to adhere to the schema. In this example, LLM simply copies the text ("affordable") from the turn as a DST prediction, while SLM is capable of grounding the value in the schema-specific format ("cheap"). However, we identify two strengths that LLM possesses over SLM. Firstly, it excels in handling common-sense knowledge, for example, it can infer the correct number of guests staying at the guest house from the context ("me and my mum"). Secondly, it demonstrates proficiency in long-context reasoning. When there is a reference to previous context across domains, LLM consistently makes the correct inference, while SLM often overlooks the context and produces random values.

\input{tables/newlm}

\subsection{Routing a New LM}
To demonstrate the flexibility of \ourmodelshort, we provide routing results of when integrating a new LM, specifically T5-3b finetuned with Prompt-DST method on MultiWOZ 5\% training data. We apply OrchestraLLM with an off-the-shelf SenBERT to route between T5-base and T5-3b. The results displayed in Table~\ref{tab:newlm} underscore the adaptability of \ourmodelshort in effectively routing examples with a newly introduced LM, all without requiring any additional training of the retriever.

%% file: tables/example.tex
\begin{table*}[th]
    \small
    \centering

    \vspace{-0.5\baselineskip}
    \begin{tabular}{l|p{13cm}}
        \toprule
        \multicolumn{2}{c}{\textbf{Example from SLM pool}} \\
        \midrule
        \colorbox{mintgreen}{\textbf{DST of Previous Turn}} & \textit{restaurant-area: centre} \\
        \colorbox{lavenderpurple}{\textbf{Test turn}} &  [system] Do you have a cuisine or price range in mind? [user] Yes, something in the affordable price range. Also, do any of them serve Singaporean food ?\\
        \colorbox{oceanblue}{\textbf{SLM Prediction}} & \textit{restaurant-food: Singaporean, restaurant-pricerange: cheap} \\
        \colorbox{sunsetorange}{\textbf{LLM Prediction}} & \textit{restaurant-food: Singaporean, \textcolor{red}{restaurant-pricerange: affordable}} \\
        
        \midrule
        
        \multicolumn{2}{c}{\textbf{Example from LLM Pool}} \\
        \midrule
         \colorbox{mintgreen}{\textbf{DST of Previous Turn}} & \textit{hotel-name=Alpha Milton guest house} \\
        \colorbox{lavenderpurple}{\textbf{Test turn}} &  [system] Would you like to book a room? [user] That would be a massive help if you can do that for me! It's me and my mum and we'll be there for 2 nights.\\
       \colorbox{oceanblue}{\textbf{SLM Prediction}} & \textit{hotel-bookstay: 2, \textcolor{red}{hotel-bookpeople: 1}} \\
       \colorbox{sunsetorange}{\textbf{LLM Prediction}}  & \textit{hotel-bookstay: 2, hotel-bookpeople: 2} \\
        
        \midrule
        
         \colorbox{mintgreen}{\textbf{DST of Previous Turn}} & \textit{restaurant-name=Cocum, restaurant-area: west} \\
        \colorbox{lavenderpurple}{\textbf{Test turn}} &  [system] Can I be of any further assistance today? [user] Yes, I am also looking for a 3-star hotel located in the same area as the restaurant.\\
        \colorbox{oceanblue}{\textbf{SLM Prediction}} & \textit{hotel-stars: 3, \textcolor{red}{hotel-area: centre}} \\
        \colorbox{sunsetorange}{\textbf{LLM Prediction}}  & \textit{hotel-stars: 3, hotel-area: west} \\
        \bottomrule
    \end{tabular}
        \caption{Representative examples from SLM and LLM pool. \textcolor{red}{Red color text} indicates the errors made by LMs.}
    \label{tab:example}
\end{table*}

%% file: tables/newlm.tex
\begin{table}[t]
    \small
    \centering
    \begin{tabular}{l|cc}
    \toprule
        \textbf{Models}  & \textbf{\textit{Assignment}}  & \textbf{TLB} \\
        \midrule
       T5-base & N/A  & 73.43    \\
       T5-3B &  N/A   & 78.77   \\
       \ourmodelshort (T5-base+T5-3B)  & 61\% & 81.09  \\
        \bottomrule
    \end{tabular}
    \caption{Routing results between T5-base and T5-3B(a new LM) using an off-the-shelf SenBERT. We denote the \% of testing turns routed to T5-base as \textit{Assignment}.}
    \label{tab:newlm}
\end{table}

%% file: sections/related2.tex
\section{Related Work}



\subsection{Sample-Adaptive Inference}
For allocating variable levels of computational resources for processing different input samples, two predominant categories of approaches have emerged in this domain: early exiting~\cite{liu-etal-2020-fastbert, xin-etal-2021-berxit, zhou2020bert} and token dropping~\cite{goyal2020power, guan2022transkimmer, kim-cho-2021-length}. ~\citet{salehi2023sharcs} also studies to direct different samples to sub-networks with varying
widths. Our proposed routing framework also embraces sample-adaptive inference. However, it distinguishes itself by leveraging not just a single model but a combination of models.
\vspace{-0.7mm}

\subsection{Model Switching}
There has been a growing body of research focusing on the concept of model switching, wherein input examples are intelligently routed between small and large models based on their individual complexity levels. For instance, ~\citet{madaan2023automix} proposes a methodology that leverages an external meta-verifier to ascertain the correctness of predictions made by a SLM and to determine whether an example warrants routing to a LLM. In contrast, our approach does not necessitate the use of additional verifiers. Another set of related approaches, exemplified by the work of \citet{vsakota2023fly,kag2022efficient}, involves training binary classifiers to categorize examples as suitable for SLM or LLM processing. This approach requires the router to be trained on labeled data where language models have made predictions. In contrast, our methodology exhibits the ability to leverage off-the-shelf retriever, enhancing its versatility.

\subsection{Few-Shot Dialogue state tracking}
To reduce the need for labeled data in DST, many approaches are proposed for few-shot DST~\cite{li-etal-2021-zero,lin2021leveraging,shin-etal-2022-dialogue,hu-etal-2022-context}. 
The state-of-the-art few-shot DST model is~\cite{king-flanigan-2023-diverse}, in which the authors reformulate DST as a Python programming task and leverages Codex~\cite{chen2021evaluating} as the backbone LLM, which is no longer accessible. Additionally, their approach involves multiple decoding passes for a single turn and relies on probability scores of tokens, which might not always be readily available.

%% file: sections/conclusion.tex
\section{Conclusion}
We introduce \ourmodelshort, a routing framework that seamlessly integrates a SLM and a LLM, orchestrated by a retrieval-based router. During inference, a dynamic router guides instances to either LM based on their semantic embedding distances with the retrieved LM exemplars, leveraging the expertise of both SLM and LLM. Our evaluation on DST demonstrates that \ourmodelshort outperforms LLM-based systems while also achieving computational cost savings of over 50\%. This research represents a significant step towards efficient collaboration of language models, particularly in a multi-turn human-computer interaction system such as task-oriented dialogue.

%% file: sections/limitations.tex
\section{Limitations}
Our study demonstrates the benefits of combining SLM and LLM for improved task performance while managing computational costs, particularly in the context of dialogue state tracking tasks. However, it's important to acknowledge that the applicability of our approach may not extend seamlessly to all types of tasks in the broader NLP domain. Additionally, in our current framework, we focus on leveraging a SLM and a LLM. However, real-world applications often involve a wide array of diverse tasks, each potentially requiring LMs with varying expertise. As a future avenue of research, we intend to explore the orchestration of multiple LMs simultaneously.

%% file: sections/appendix.tex
\pagebreak
\section{Related Work}
\label{appendix:related_work}
\subsection{Sample-Adaptive Inference}
To enable variable levels of computational resources to be allocated for processing different input samples, two predominant categories of approaches have emerged in this domain: early exiting and token dropping. Early exiting strategies typically incorporate additional classifiers at intermediate layers within a model, determining whether a given input example should terminate its processing prematurely and abstain from propagating to subsequent layers~\cite{liu-etal-2020-fastbert, xin-etal-2021-berxit, zhou2020bert}. On the other hand, token-dropping techniques aim to dynamically reduce the length of the input sequence, achieved by selectively excluding certain tokens from the input sequence, which are subsequently not passed on to subsequent layers in the model~\cite{goyal2020power, guan2022transkimmer, kim-cho-2021-length}. ~\citet{salehi2023sharcs} also studies to direct different samples to sub-networks with varying
widths. Our proposed routing framework also embraces sample-adaptive inference. However, it distinguishes itself by leveraging not just a single model but a combination of models.

\subsection{Model Compression}
Model compression primarily falls into three major paradigms: pruning, distillation, and quantization. Pruning strategies are primarily devised to reduce computational overhead by selecting and retaining a subnetwork within a larger model~\cite{fan2019reducing, michel2019sixteen, wang2020structured, lagunas2021block, xia2022structured}. In contrast, distillation techniques entail the training of a compact student model, with the objective of imparting the knowledge and performance of a larger teacher model~\cite{sanh2019distilbert, turc2019well, jiao2020tinybert}. Finally, quantization methods aim to diminish memory demands by representing model parameters with fewer bits, thereby trading off a degree of precision for enhanced efficiency~\cite{shen2020q, dettmers2022llm, dettmers2023qlora}. Note that the aforementioned methods are characterized as "static" in nature, as they primarily focus on the optimization of fixed model architectures for each data point. In contrast, the routing framework introduced in this work adopts a dynamic perspective.